\documentclass{article}
\usepackage{mathtools}
\usepackage{amsmath}
\usepackage{times}
\usepackage{graphicx}
\usepackage{subfigure} 

\usepackage{natbib}

\usepackage{algorithm}
\usepackage{algorithmic}
\usepackage{amsmath}
\usepackage{amssymb}

\usepackage{hyperref}

\usepackage[accepted]{icml2017}

\icmltitlerunning{A Self-Attention Network for Hierarchical Data Structures}

\begin{document} 

\twocolumn[
\icmltitle{A Self-Attention Network for Hierarchical Data Structures with an Application to Claims Management}
\begin{icmlauthorlist}
\icmlauthor{Leander Loew}{to}
\icmlauthor{Martin Spindler}{goo}
\icmlauthor{Eike Brechmann}{to}
\end{icmlauthorlist}
\icmlaffiliation{to}{AI Developments, Allianz, Germany}
\icmlaffiliation{goo}{University of Hamburg, Germany}
\icmlcorrespondingauthor{Leander Loew}{leander.low@outlook.de}
\vskip 0.3in
]

\printAffiliationsAndNotice{}
\graphicspath{{images/}}

\begin{abstract} \

Insurance companies must manage millions of claims per year.
While most of these claims are non-fraudulent, fraud detection is core for insurance companies.
The ultimate goal is a predictive model to single out the fraudulent claims and pay out the non-fraudulent ones immediately.
Modern machine learning methods are well suited for this kind of problem.
Health care claims often have a data structure that is hierarchical and of variable length.
We propose one model based on piecewise feed forward neural networks (deep learning) and another model based on self-attention neural networks for the task of claim management.
We show that the proposed methods outperform bag-of-words based models, hand designed features, and models based on convolutional neural networks, on a data set of two million health care claims.
The proposed self-attention method performs the best.

\end{abstract} 

\section{Introduction}
\label{Introduction }

Under private insurance, health care costs must usually be paid up front by the insuree to the health care provider (doctor, hospital, etc.).
In return, the insuree gets a claim which is then handed in to the insurance company for reimbursement.
The insurance company now faces the problem of either paying out the claim completely or determining if a ``correction'' is necessary (overcharged or fraudulent bill).
This is the so-called claim management problem.

The data set consists of past claims and their associated labels.
Each claim is classified either as correct or as having had a ``correction'' made to it (``fraudulent claim'').
Modern supervised machine learning methods are tailored for such problems.
Given a claim, we want to predict the probability of fraud and pay out all non-fraudulent claims immediately.

However, health care claims have an unusual input data structure.
Each claim consists of multiple claim rows.
E.g. at each consultation of a doctor or in a hospital, several operations are conducted and usually each operation is billed according to some prescribed compensation scheme, which is obligatory.
Those operations, coded, form the rows of a claim.
The number of operations varies from patient to patient, resulting in claims of varied lengths.
Each claim row consists mainly of three variables: A procedure code, a factor, and the costs of this treatment (as a numerical value).
The procedure code encodes the treatment received.
For each procedure code there is a prescribed basis amount, which is multiplied by the factor, yielding the costs of this treatment.
Each triple, of procedure code, factor, and numerical amount, in one row, we call a ``claim row''.
We can think of the input data structure per observation as a matrix with a variable number of rows.

Since ordinary machine learning methods (like gradient boosting or random forests) require a fixed size vector as input, the standard procedure has been to find a fixed size vector by manual feature engineering based on domain knowledge.
This feature engineering requires costly domain experts, does not scale to similar problems well, and good features can be very hard to find.

In search of a better solution and a way to automate the feature engineering, we exploit the similarities between the data structure of a claim and the data structure of a text: Both texts and claims consist of a sequence of vectors for each observation: in the case of a text, these are of words; in the case of a claim, these are rows.

\begin{figure}[ht]
	\vskip 0.2in
	\begin{center}
		\centerline{\includegraphics[width=\columnwidth]{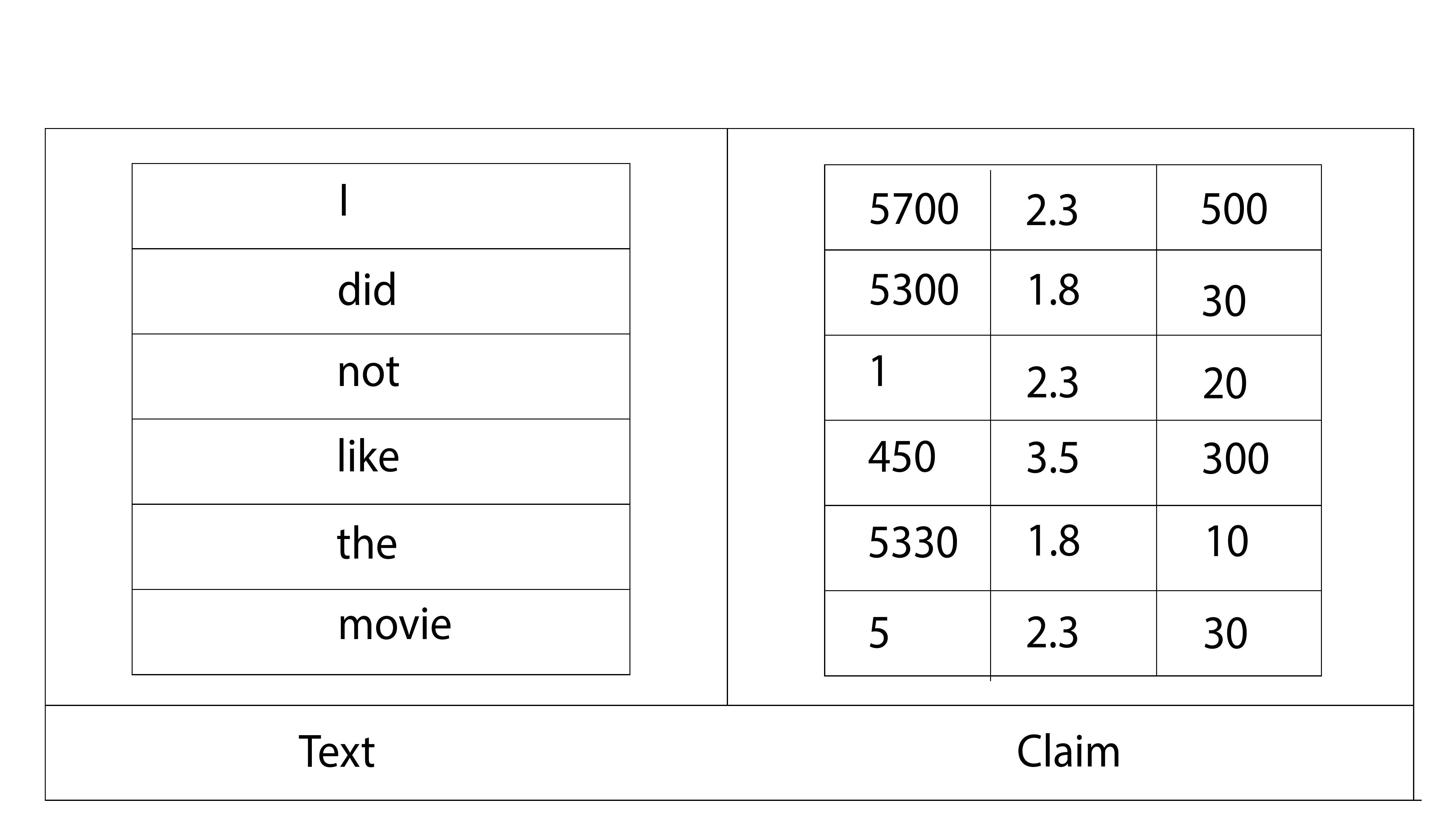}}
		\caption{Comparison between text input and claim input for a single observation.
Both consist of a sequence:
for texts, a sequence of one variable -- words.
For claims, a sequence of 3 variables: procedure code, factors, and numerical amounts.
Each triple (procedure code, factor, numerical amount) in one row will be referred to as a ``claim row''.
In contrast to words, the ordering of ``claim rows'' is arbitrary.
}
		\label{icml-historical}
	\end{center}
	\vskip -0.2in
\end{figure} 

In recent years, machine learning methods based on artificial neural networks have started to show state of the art performance in the domain of text processing.
Neural network-based methods can handle unusual data structures directly and feature engineering as part of the learning process.
Thus, they may be an automated alternative to manual feature engineering for unusual data structures, such as claims.
We pursued this approach also for the claim management problem.

\section{Sequence Classification}

In a sequence classification problem, as defined in \cite{Graves2012}, the task is to predict a fixed size vector, usually a probability vector, a sequence of real numbers.
In this way, the claim management problem can be seen as a sequence classification problem: The input is a sequence of claim rows and the task is to predict the scalar probability of fraud.

A well-studied sequence classification task is sentiment analysis.
Here the task is -- given a sequence of words -- to predict the “sentiment” of the sentence.
This is usually the scalar probability of a negative or positive sentiment.
Most machine learning methods for sentiment analysis either rely on a bag-of-words model of words or on neural network based methods.

\textit{Bag-of-Word Models}

In the bag-of-words type of model, we find a fixed size representation of the words by one hot encoding them and summing them up.
Afterwards, machine learning methods, e.g. a feed forward network, can be applied.
Bag-of-words type models are also applicable to claim management and will be used as the baseline model.

\textit{Neural Network based Models}

Neural network based methods for sentiment analysis usually rely on four distinct types of layers: 
The first type is an embedding layer, to turn the sequence of words into a sequence of real vectors.
One can either use pre-trained word embeddings, as in \cite{Mikolov2013}, or learn the embeddings as part of the training process.

The second type is a feature extraction layer, which turns the sequence of inputs into a sequence of context dependent representations.
Most sequence classification models either rely on recurrent neural networks (RNN) or convolutional neural networks (CNN) \cite{Kim2014}.

The third type is an aggregation layer, which is used to turn the sequence of feature vectors into one fixed size vector; most often max pooling or neural attention \cite{Bahdanau2014} are used.

The fourth type is a fully connected layer, to get a final prediction from the aggregated feature representation.
We use a feed forward network here to be able to train the whole structure end to end.

We will follow this structure, viz. embedding, feature extraction, aggregation, and fully connected layer, but there are some important differences between claim and text classification that necessitate a different feature extractor, as we will describe in the next chapter.

\subsection{Comparison between text and claim classification}
Exploiting the similarity in data structures, it seems natural to transfer  models developed for sentiment analysis to the task of claim management.
However, although words in a text as inputs have a meaningful ordering, claim rows in a claim do not.

This has important consequences for the feature extraction layer.
Recall that the task of the feature extraction layer is to derive a context dependent representation, i.e. the “meaning”, of an element of the input sequence.
To understand the meaning of a word, the feature extraction mechanism should incorporate the ordering of the words.

This makes recurrent neural networks particularly useful for text processing.
RNNs can exploit the ordering of the words and the locality present in texts.
Another method applicable for text processing is that of convolutional neural networks.
They do not maintain the ordering of the sequence and are only able to express local dependency.

Due to the random ordering of the claim rows, we need a model which is invariant with respect to the ordering of the sequence but at the same time able to form context dependent representations.
Both CNNs and RNNs seem inappropriate for this task.
Instead, we propose three models: 1) a model which is very similar to the classical bag-of-words model,
2) a model based on piecewise feed forward neural networks, and 3) a model based on self-attention neural networks.

\begin{figure}[ht]
	\vskip 0.2in
	\begin{center}
		\centerline{\includegraphics[width=\columnwidth]{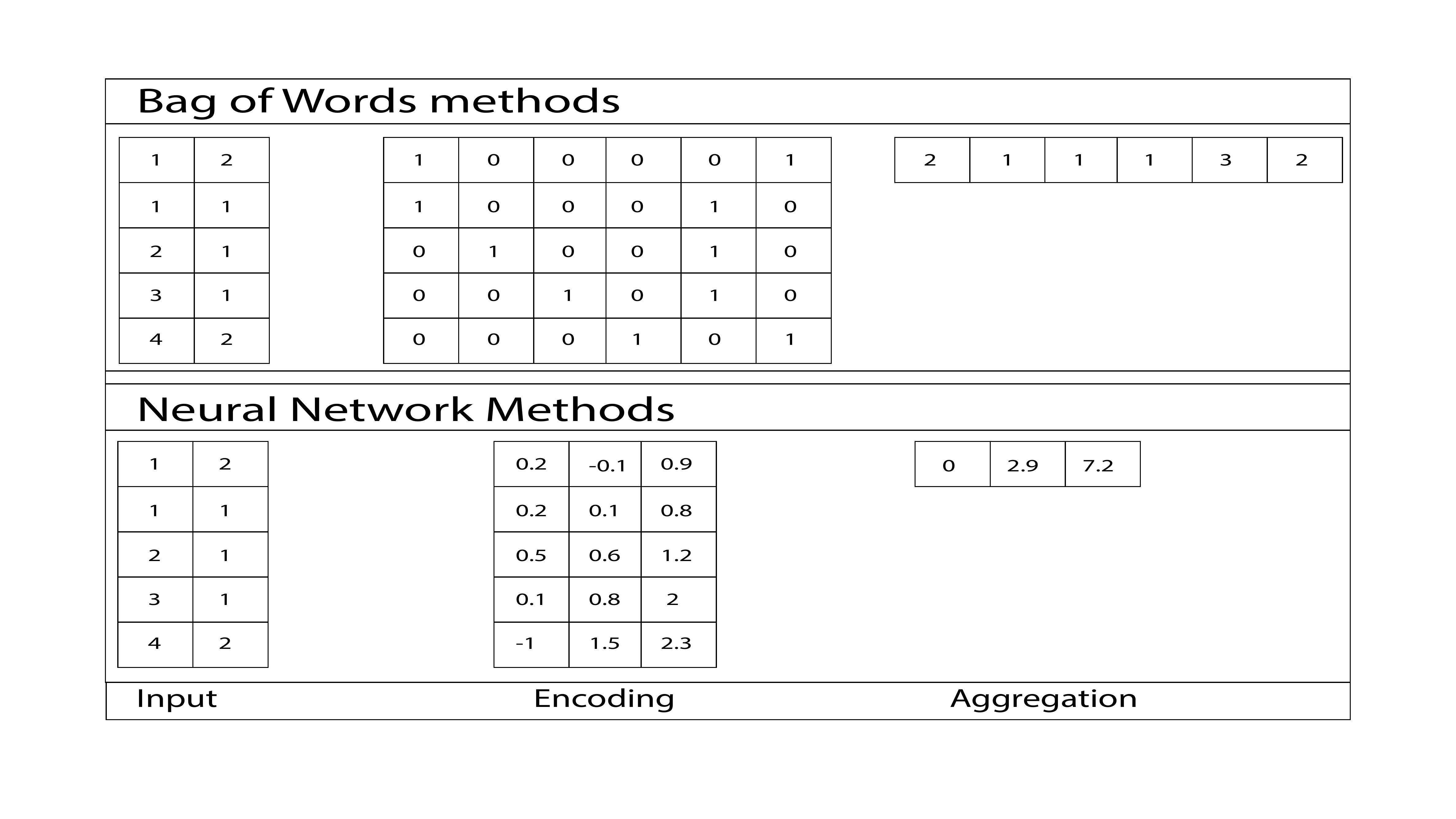}}
		\caption{Comparison between bag-of-words type methods and neural network based methods.
The example has only two categorical variables, one with four and the other with two levels.
For the bag-of-words method, we one hot encode the categorical variables and then sum them up.
For neural network based methods we ``encode'' the input variables by an embedding followed by a feature extraction layer, which results in a dense representation.
Afterwards we apply the aggregation in this dense space, here by summing.
}
		\label{icml-historical}
	\end{center}
	\vskip -0.2in
\end{figure}

\subsection{Bag-of-Words model}
As the baseline model we use a bag-of-words type model, where we one hot encode the categorical variables and sum them up.
Similarly, we form the sum of the numerical variable.
This model has the problem that we lose the association between variables which constitute a claim row.

\subsection{Piecewise feed forward model}
In the context of sequence classification, a piecewise feed forward network can be thought of as a convolutional neural network with a single claim row as context.
Unlike the bag-of-words model, the piecewise feed forward model keeps the relations between the variables on the same claim row.

A drawback of the piecewise feed forward model is that it does not form a ``context dependent representation'', i.e. the triple of procedure code, factor, and amount will always have the same kind of representation.

To best understand this, we can compare it to texts: In texts, words are almost completely defined by their context.
We can see this clearly with words that have multiple meanings.
E.g. the word \textquotedblleft nails\textquotedblright\ can either mean finger nails or nails made out of metal.
Similarly, negation plays a big role.
The word \textquotedblleft like\textquotedblright has the exact opposite meaning if the word \textquotedblleft not\textquotedblright\ came previously to it.
From these two examples, we can see that context plays a very important role in the interpretation of language, and thus it is very useful to find context dependent representations.
We conjecture that a context dependent representation might be advantageous for the interpretation of a claim row.
 
\begin{figure}[ht]
\vskip 0.2in
\begin{center}
\centerline{\includegraphics[width=\columnwidth]{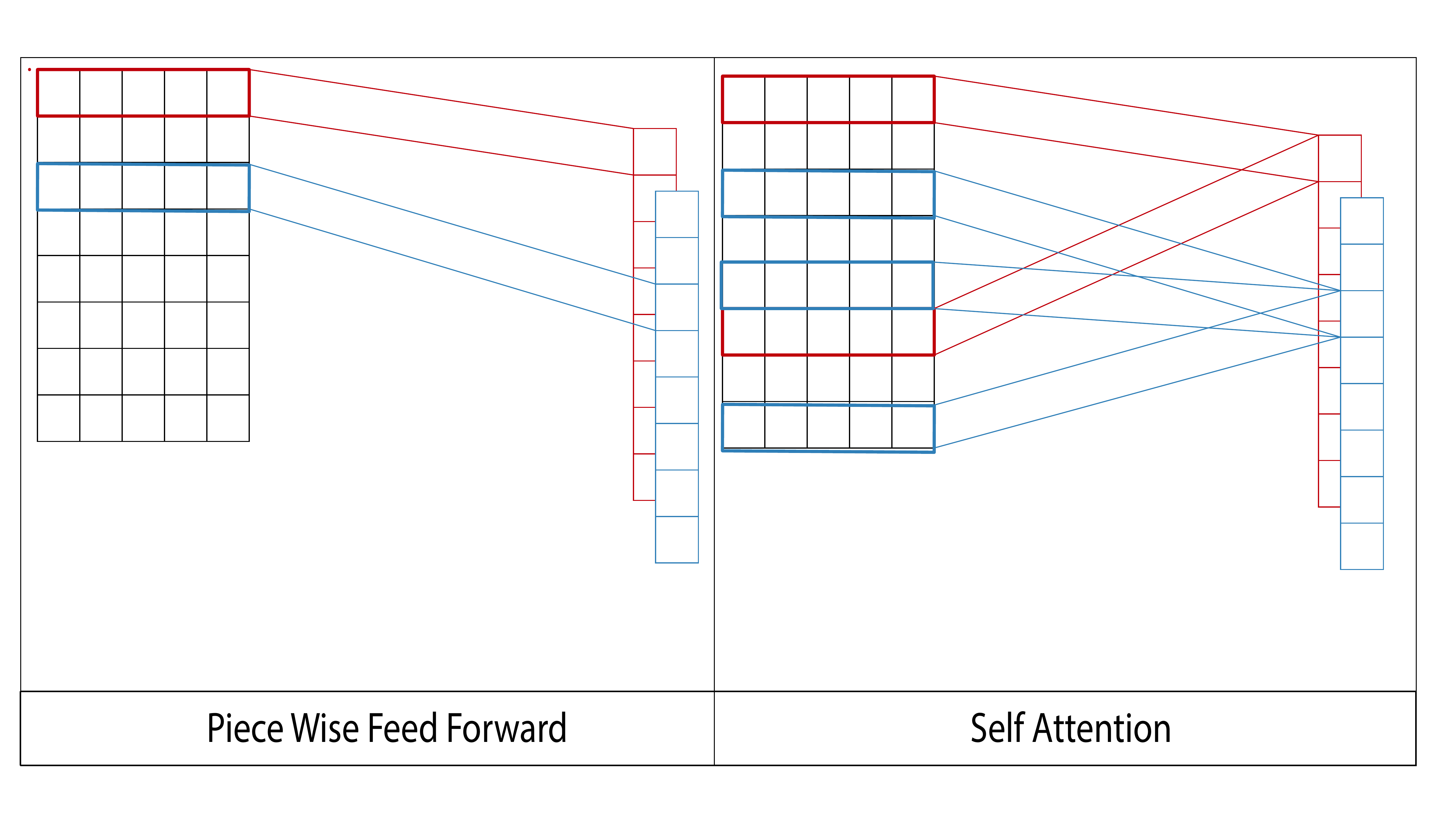}}
\caption{Comparison between piecewise feed forward and self-attention feature extractors.
For piecewise networks, a feature of an input element depends only on its input.
But in self-attention networks, a feature of an element of the sequence will depend on both itself and on any other elements of the sequence.
How much it depends on the other elements is defined by the attention weights.
This way, the self-attention mechanism can form a context dependent representation of a claim row.}
\label{icml-historical}
\end{center}
\vskip -0.2in
\end{figure}

\subsection{Self-Attention model}

Self-Attention was introduced in \cite{Vaswani2017} where it was applied to the task of translation, and in \cite{Shen2017} to the task of sequence classification.
The main benefit of self-attention networks in comparison to piecewise feed forward networks is that they can form a context dependent representation, as described in the previous chapter.

The intuition of self-attention is as follows. For each input $i$, we define an attention distribution over the other inputs.
This attention distribution will give high weight to the inputs $j\neq i$ which are relevant for the interpretation of the input $i$.
We then use those attention weights in combination with the other inputs $j \neq i$ to form the derived feature for the input $i$.

Thus, the derived features of an input can depend on any other element of the input sequence, independently of their placement.
We can even analyse the attention weights to find out which other inputs were important for the interpretation of a particular input.
So, for self-attention, the ordering of the sequence is irrelevant.
It can form a context dependent representation and it is even highly interpretable.
Accordingly, self-attention seems valuable for the application to claim classification.

\section{Description of the Model} 
 
We will next describe the neural network used for the task of claim classification in detail.
The model for this application consists of four distinct types of layers: an embedding layer for the categorical variables, a feature extraction layer to derive features from the sequence of inputs, an aggregation layer to find a fixed size representation, and a fully connected layer to form a scalar prediction.

\begin{figure}[ht]
\vskip 0.2in
\begin{center}
\centerline{\includegraphics[width=\columnwidth]{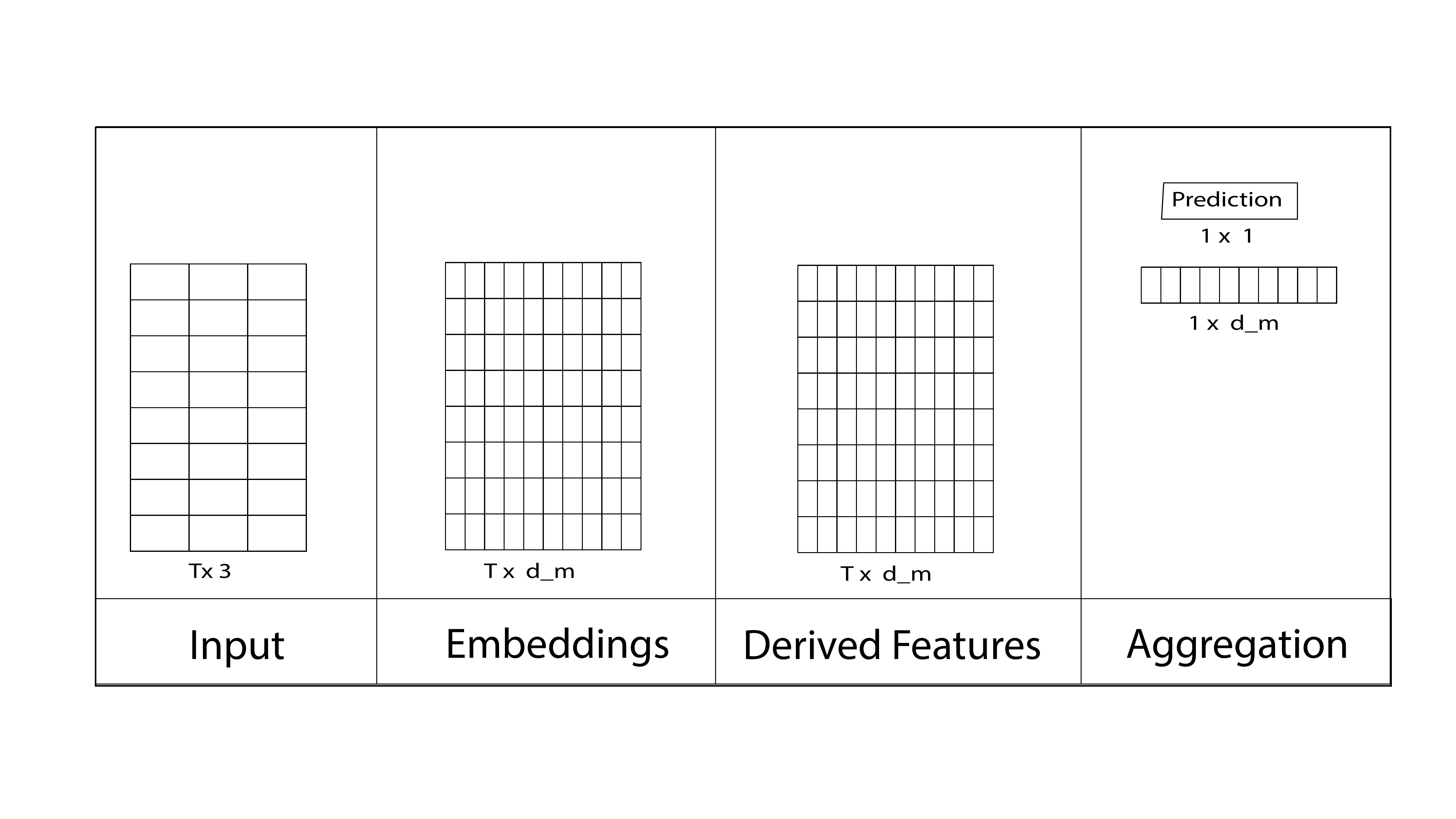}}
\caption{Overview of the model.
}
\caption{A sequence of categorical and numerical variables is turned into a sequence of real vectors by embedding the categorical variables and then applying a feature extraction layer to derive a sequence of features.
This sequence of features is turned into a fixed size vector by an aggregation layer.
Lastly, we use a fully connected layer to derive a scalar prediction.
}

\label{icml-historical}
\end{center}
\vskip -0.2in
\end{figure} 

\subsection{The embedding layer}

Categorical embeddings are an adaptation of word embeddings to general categorical variables, as in \cite{Guo2016}.
Categorical embeddings associate a dense representation to each input level, learned during the training process.
The embedding layer is especially useful for this problem, due to the high cardinality of one of our categorical variables.
When one-hot encoding the categorical variables, this high cardinality leads to too many parameters and thus can quickly lead to overfitting.
Replacing the one-hot encoding with an embedding layer reduces the number of parameters substantially, increases the training and inference time, and leads to better results.

We will represent the output of the concatenation of the embedding layer as our input matrix $X$, of dimension $d_{m} \times T$,
where $T$ is the number of claim rows (analogous to the number of words or time steps) and $d_{m}$ is the size of our hidden representation, which is a hyperparameter.

\subsection{The feature extraction layer}
The task of the feature extraction layer is to turn the sequence of input variables into a sequence of low dimensional dense features.
We compare two different feature extraction layers: A piecewise feed forward network and a self-attention network.
The main difference is that the piecewise feed forward network forms a context independent representation, while the representation formed by self-attention can incorporate the context.

\textit{Piecewise Feed Forward}

For each observation, the sequence of inputs represented by the matrix 
$X$ gets turned into a sequence of features $H$ by applying a piecewise feed forward model.
\begin{center}
	$H= relu(W_1X+b_1)$
	
	$ W_1 \in \mathbb{R}^{d_{m} \times d_{m}}; H \in \mathbb{R}^{d_{m} \times T} ;$	
	$b_1 \in \mathbb{R}^{d_{m}}$
\end{center}

\textit{Self-Attention}

For each observation, the sequence of inputs represented by the matrix 
$X$ is turned into a sequence of features $H$ by applying scaled dot product self-attention, as in \cite{Vaswani2017}, who we follow.
Here we use a self-attention model with a single head and a single layer.
We first derive a set $K$ of keys, a set $V$ of values, and a set $Q$ of queries, from linear projections of our inputs,
a matrix multiplication by a parameter matrix followed by adding a bias:

\begin{center}

$Q= W_q X + b_q$

\end{center}

\begin{center}
$K= W_K X + b_k$
\end{center}

\begin{center}
$V= W_V X + b_v$
\end{center}

\begin{center}
$W_Q,W_K,W_V\in \mathbb{R}^{d_{m} \times T}$,
$b_q, b_k, b_v \in \mathbb{R}^{d_{m}}$,
$Q, K, V \in \mathbb{R}^{d_{m} \times T}$.
\end{center}

Then we use the scaled dot product attention operation to turn the keys, queries and values into a sequence of features $H$: 

\begin{center}
$H_1=softmax(\frac{QK^T}{\sqrt{d_m}})V$

$ H_1 \in \mathbb{R}^{d_{m} \times T} $
\end{center}

Afterwards a residual and layer normalization layer as in \cite{Wu2017} and \cite{Ba2016} is applied.
\begin{center}
$H_2 =layernorm(H_1+X)$ 
\end{center}

Next we apply a piecewise feed forward layer to these derived features.
\begin{center}
$H_3= W_2relu(W_1H_2+b_1)+b_2$

$ W_1 \in \mathbb{R}^{d_{m}*2 \times d_{m}};W_2 \in \mathbb{R}^{d_{m} \times d_{m}*2}; H \in \mathbb{R}^{d_{m} \times T} $
\end{center}
$b_1$ and $b_2$ are biases of dimension $d_m$.

Lastly we apply another normalization and residual layer: 
\begin{center}
$H_4=layernorm(H_3+H_2)$
\end{center}

The self-attention algorithm is summarized in Algorithm 1.

\begin{algorithm}[tb]
\caption{Self-Attention}	
\begin{algorithmic}
\STATE {\bfseries Input:} sequence X
\STATE {\bfseries Query, Key, Value:} Linear(X) 
\STATE {\bfseries $H_1$:} Attention(Query, Key, Value) 
\STATE {\bfseries $H_2$:} Residual+Normalization($H_1+X$) 
\STATE {\bfseries $H_3$:} Pice Wise Feed Forward($H_2$) 
\STATE {\bfseries $H_4$:} Residual+Normalization($H_2+H_3$) 
\STATE {\bfseries Output:} $H_4$

\end{algorithmic}
\end{algorithm}

The features derived from self-attention can form context dependent features and those features are by design highly interpretable.
We can see which other elements in the input sequence were important for the features of a particular row.
It is crucial that all parameters are independent of the sequence length $T$.
This is a requisite for incorporating variable sequence lengths and generalizing even to unseen lengths.

\subsection{The aggregation layer}
To efficiently incorporate variable length sequences, we must find one fixed size vector for each possible sequence length.
This is the task of the aggregation layer.

We start out with a sequence of features which we represent by the matrix 
\begin{center}
$H \in \mathbb{R}^{d_{m} \times T}$.
\end{center}
The task of the pooling layer is to find a fixed size representation.
\begin{center}
$h \in \mathbb{R}^{d_{m}}$
\end{center}

Sum, mean, and max pooling are simple methods which either sum, take the average, or the maximum value over the corresponding dimension of the feature tensor.
Attention is a more advanced way and can be described as taking a weighted average, where the weights are learned.
For claims processing, we observe a positive correlation of fraudulent claims with the sequence length.
Thus,  our aggregation method should be able to naturally scale with the sequence length. For this reason, sum pooling seemed like the obvious choice.
However, we modified sum pooling slightly, and call this form of pooling “sigmoid attention”.

First we derive the attention weights as a linear layer with sigmoid activation:
\begin{center}
	$a= sigmoid(W_a H + b_a)$
\end{center}
\begin{center}
$H \in \mathbb{R}^{d_{m} \times T};$
$W_a \in \mathbb{R}^{T};$
$b_a \in \mathbb{R}^{d_{m}};$
$a \in \mathbb{R}^{T};$
\end{center}
Next we use these weights to form a weighted sum of the elements of the feature sequence to derive a fixed size vector.
\begin{center}
	$h=  Ha$
	
	$h \in \mathbb{R}^{d_{m}}$
\end{center}

In contrast to attention with the softmax operator, our attention weights are generated by the sigmoid operator with weights not summing to one.
Similar to ordinary attention, this allows a high degree of interpretability, because for each observation we can interpret the attention weights to identify important elements of the feature sequence.

\subsection{The feed forward layer}

After the aggregation layer we have one vector per observation and thus essentially a “normal” machine learning problem. To be able to train the whole architecture end to end, we set up a feed forward network on top.

\section{Empirical Evaluation}

\subsection{The data}
The empirical study employed
two million health care claims for training, testing, validation, and comparison. 
We used a validation set for model selection and an independent test set, both of size 400,000.
The final results for the test set are presented. Convergence images were made from the validation data.
For each observation, we have a sequence of input vectors ranging from 1 to 100 input vectors.
Each input vector consists of three variables.
The first is a numerical variable, which we scale between zero and one after log transforming.

The second variable is a factor variable with six levels.
The procedure code is a categorical variable with over $4,000$ levels.
We can visualize each input as a two-dimensional matrix, where each row is one claim row consisting of the three variables.

Each data point has an associated label: If there was a correction amount or if everything was fine with the claim.
For example, if the original total numerical amount was 100 Euro and we paid out only 80, we call the difference the correction amount.
We call the existence of a correction amount a ``fraud'' case and code it with a 1. A zero is a non-fraudulent case.
The response is highly imbalanced.
We also observe the actual numerical correction amount, which we will use to scale the loss function.

\subsection{Performance metric}
Because of the imbalanced response, standard evaluation metrics like accuracy and cross entropy loss are not useful.
The default metrics for imbalanced problems are the area under the receiver operator curve and the area under precision recall curve.
However, for this specific problem, we derived an additional metric, as it turns out for each claim we have a very quantifiable cost of misclassification:
If we have a claim which is unproblematic, and we classify it as problematic, we have the cost of the clerk who must look at the claim, which we can estimate and assume fixed for all claims: $k$.
If we have a claim which is problematic, and we classify it as unproblematic, we have the cost of not receiving the correction amount $c_i$.
We use those derived misclassification costs for scaling the loss function appropriately.
Furthermore, we can define a profit metric for model selection.
First we define the benefit of our model.
This is the total payments resulting from the true positives.

\begin{center}
$Benefit = \sum_{i=1}^{N} I(P(x_i)>Threshhold)*(c_i-k)$
\end{center}

Then we define the cost of our model.
This is the total cost of the false positives.
\begin{center}
$Cost =  \sum_{j=1}^{M} I(P(y_j)>Threshhold)*k $
\end{center}

\noindent Here, $N$ is the number of fraudulent cases,  $M$ the number of non-fraudulent cases, and $x_i$ and $y_j$ are the members of the corresponding classes.
$P()$ is the predicted probability given by our machine learning model.
$I()$ is the indicator function.
$Threshhold$ is usually set to be $0.5$ and $c_i$ is the correction amount associated with a particular case, which represents the benefit for that case if correctly classified.
$k$ is the fixed cost of claim processing if it is processed manually.

We then define the potential, which is just the maximal possible profit divided by a normalizing factor.

\begin{center}
$Potential  = \sum_{i=1}^{N} (c_i-k)$
\end{center}

Lastly, our final profit metric is benefit minus cost divided by the normalizing factor.

\begin{center}
$Profit= \large \frac{Benefit  -  Cost}{ Potential }$
\end{center}
\normalsize

For $k$ we used an internal estimate.
However, we observed that the relative performance was not dependent on this estimate.
The main benefit of the loss scaling is to differentiate the positive cases from each other instead of differentiating the zero cases from the one cases.

\subsection{Training}

For all models, we used the Adam optimizer with learning rate 0.0001 and weight decay of 0.00001 and early stopping and validation loss as the stopping metric.
We selected the hyperparameters by random search. The selected hyperparameters are presented in Table 1.
As our loss, we used a scaled cross entropy loss function, where the sample weights for loss scaling are the correction amounts divided by a normalizing factor.
We trained on batches of size 128 with claims of the same size.
The model has been implemented in Keras.

\begin{table}[t]
	\caption{ Hyperparameters per Model.}
	\label{sample-table2}
	\vskip 0.15in
	\begin{center}
		\begin{small}
			\begin{sc}
				\begin{tabular}{lcccr}
					\hline
					\abovespace\belowspace
					Model & Dim FE & Dim FC & Dropout & Weight Decay \\
					\hline
					\abovespace
					CNN & 64 & 512 & 0 & $1e07$\\
					BOW & - & 1024& 0.1 & $1e05$ \\
					PFF  & 32 & 512& 0 & $1e06$ \\
					SelfA &  128  & 512 & 0 & $1e05$ \\

					\hline
				
				\end{tabular}
		
			\end{sc}
		\end{small}
	\end{center}
	\vskip -0.1in
\end{table}

\subsection{Model comparison}
We use the area under the receiver operator curve, the area under the precision recall curve, and our defined profit as our valuation metrics.
All metrics are reported on a test set.
Results are reported in Table 1.
As we can see, the self-attention model performs the best.
We observe a big jump between the performance of CNN, bag-of-words, hand designed features, and our preferred methods: PFF and self-attention.
We explain this difference in performance by the fact that that self-attention and piecewise feed forward networks keep the relation between variables on the same claim row intact, while all other methods lose this relation.
From this we infer that the interactions between variables on the same claim row are important.

Furthermore, the self-attention model is capable of forming a context dependent representation of a claim row, which explains the benefit in performance in comparison to the piecewise feed forward model.

\begin{figure}[h!t!]
	\vskip 0.2in
	\begin{center}
		\centerline{\includegraphics[width=\columnwidth]{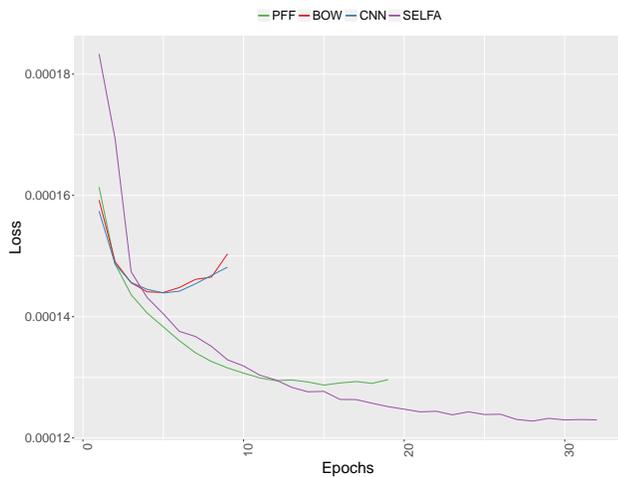}}
		\caption{Validation loss convergence plots of the different methods.
We can see that CNN and BOW models quickly start overfitting, while the PFF and self-attention model reach much lower loss values, with self-attention performing the best.
We used early stopping for all methods, which explains the difference in training epochs.
 }
		\label{icml-historical}
	\end{center}
	\vskip -0.2in
\end{figure} 

\begin{table}[t]
	\caption{Model Comparison.}
	\label{sample-table}
	\vskip 0.15in
	\begin{center}
		\begin{small}
			\begin{sc}
				\begin{tabular}{lcccr}
					\hline
					\abovespace\belowspace
					Model & AUROC & AUPR & Profit \\
					\hline
					\abovespace
					CNN & 0.902 & 0.195& 0.661\\
					Manual + GBM & 0.912 & 0.231& 0.68\\
					BOW & 0.911 & 0.227& 0.673\\
					PFF    & 0.923& 0.251& 0.713 \\
					SelfA    & 0.926& 0.267 & 0.736 \\

					\hline
				\end{tabular}
			\end{sc}
		\end{small}
	\end{center}
	\vskip -0.1in
\end{table}

\section{Summary}

In this paper we analysed claims data a with deep neural nets -- a modern application of machine learning method outside of natural language processing and image analysis and with real and quantifiable economic value.
We proposed an architecture that is tailor-made for the structure of claims data.
There are many industry problems similar to claim management with hierarchical data structures, where neural network based methods can replace and improve upon hand designed features.
The next steps will be to incoporate additional input variables (fixed length and variable length information) to form a complete model of claim management.

\bibliography{library}
\bibliographystyle{icml2017}

\end{document}